\documentclass[runningheads]{llncs}

\usepackage{eccv}

\usepackage[table,xcdraw,dvipsnames]{xcolor}
\usepackage{eccvabbrv}

\usepackage{graphicx}
\usepackage{amsmath}
\usepackage{amssymb}
\usepackage{booktabs}
\usepackage{paralist}
\usepackage{bm}
\usepackage{nicefrac}
\usepackage{adjustbox}
\usepackage{bm}

\newcommand\blfootnote[1]{%
  \begingroup
  \renewcommand\thefootnote{}\footnote{#1}%
  \addtocounter{footnote}{-1}%
  \endgroup
}

\definecolor{gray}{RGB}{238, 238, 238}
\definecolor{columbiablue}{rgb}{0.61, 0.87, 1.0}

\definecolor{mygreen}{rgb}{0.4, 0.69, 0.2}
\definecolor{darkyellow}{HTML}{FFA700}
\definecolor{newpurple}{HTML}{BC61F5}
\usepackage[accsupp]{axessibility}  

\usepackage[pagebackref,breaklinks,colorlinks,citecolor=eccvblue]{hyperref}

\usepackage{orcidlink}

\begin{document}

\title{Preventing Catastrophic Forgetting through Memory Networks in Continuous Detection} 

\titlerunning{Preventing Catastrophic Forgetting through Memory Networks}

\author{Gaurav Bhatt\inst{1,2} \and
James Ross\inst{1} \and
Leonid Sigal\inst{1,2}}

\authorrunning{G.Bhatt et al.}

\institute{University of British Columbia, Vancouver, Canada \and
The Vector Institute for AI, Toronto \\
\url{https://gauravbh1010tt.github.io/}}

\maketitle

\begin{abstract}
Modern pre-trained architectures struggle to retain previous information while undergoing continuous fine-tuning on new tasks. Despite notable progress in continual classification, systems designed for complex vision tasks such as detection or segmentation still struggle to attain satisfactory performance. In this work, we introduce a memory-based detection transformer architecture to adapt a pre-trained DETR-style detector
to new tasks while preserving knowledge from previous tasks. We propose a novel localized query function for efficient information retrieval from memory units, aiming to minimize forgetting. Furthermore, we identify a fundamental challenge in continual detection referred to as {\em background relegation}. This arises when object categories from earlier tasks reappear in future tasks, potentially without labels, leading them to be implicitly treated as background. This is an inevitable issue in continual detection or segmentation. The introduced continual optimization technique effectively tackles this challenge.
Finally, we assess the performance of our proposed system on continual detection benchmarks and demonstrate that our approach surpasses the performance of existing state-of-the-art resulting in 5-7\% improvements on MS-COCO and PASCAL-VOC on the task of continual detection. Code: \url{https://github.com/GauravBh1010tt/MD-DETR}

\end{abstract}

\section{Introduction}
\label{sec:intro}


\blfootnote{Accepted in European Conference on Computer Vision, 2024 (ECCV'24)}

Despite notable achievements in computer vision, the majority of existing approaches adhere to a clean/offline learning paradigm. In this paradigm, the model is either directly trained or fine-tuned from a pre-trained generic checkpoint or foundational model on a training data split. It is then tested on a designated test split, assuming that it originated from the same underlying distribution. However, such a clean paradigm is impractical for numerous real-world applications, such as robotics, industrial systems, or personalized applications, where data may arrive in a streaming fashion with evolving distributions and tasks. A truly effective computer vision system should possess the ability to learn continuously, fine-tuning to novel tasks while preserving previously acquired capabilities.

{\em Continual (or lifelong) learning} has emerged as a prominent subfield of ML and vision that focuses on building systems with such capabilities \cite{IL_iCaRL,IL_3,IL_E2E,lwf1_li2017learning,lwf2_liu2020mnemonics,chen2018lifelong}. In the most typical scenario of class-incremental continual learning,  it is generally assumed that data arrives in tasks, and each task includes a disjoint subset of classes labeled within it \cite{IL_iCaRL,CIL1_zhou2022model,CIL2_yan2021dynamically}. A trivial approach to continual learning in this setting could take one of two forms: (1) fine-tuning a pre-trained or foundational model to each task, or (2) fine-tuning a pre-trained or foundational model continuously as tasks stream in. 
However, the former proves impractical as it leads to the proliferation of models (and consequently, memory and computational requirements) in proportion to the number of tasks. The latter is susceptible to the so-called {\em catastrophic forgetting} problem – 
model losing the ability to recognize classes encountered in earlier tasks once it is fine-tuned to the latter ones \cite{IL_iCaRL,IL_3,IL_E2E,lwf1_li2017learning,lwf2_liu2020mnemonics}. The issue of catastrophic forgetting is evident in all existing ML models, and even foundational models, like CLIP \cite{clip_forget}, struggle to retain past knowledge during fine-tuning.

To address catastrophic forgetting, many existing state-of-the-art solutions employ training samples from previous tasks \cite{IL_iCaRL,rolnick2019experience_replay,replay1_yoon2021online_coreset,replay2_tiwari2022gcr,replay3_replay_wang2022cafe}, stored in a buffer and replayed during subsequent tasks. However, the performance of such methods often declines substantially as the buffer size decreases. Replay buffer methods also fail to maintain a proper balance between stability (preserving past information) and plasticity (generalizing to current information) majorly due to disjoint fine-tuning on the buffer, which is referred to as the stability-plasticity dilemma \cite{trade_off_jung2023new,trade_org_abraham2005memory}. Moreover, the necessity of a buffer restricts their applicability in privacy-sensitive applications where storing samples from past tasks is prohibited. Alternative approaches to mitigate catastrophic forgetting include regularization \cite{ewc_huszar2017quadratic,mcdonnell2024ranpac} and parameter isolation \cite{iso_houlsby2019parameter,mallya2018packnet,iso_p_zhu2023self}. Recently, with the rise of pre-trained models, memory networks have emerged as a novel solution to reduce forgetting during continuous fine-tuning \cite{l2p_wang2022learning,wang2022dualprompt,smith2023coda,hide_prompt_wang2024hierarchical,DAP_jung2023generating}. Unlike replay buffers, memory networks eliminate the need for storing past samples; instead, a memory module is employed to allow the pre-trained model to retain pertinent information from previous tasks. The performance of memory networks relies heavily on the design of the query mechanism used to retrieve information from memory modules and a proper continual optimization strategy. 

\begin{figure*}[!t]
    \begin{subfigure}[t]{0.32\textwidth}
        \centering
        \includegraphics[width=1\linewidth]{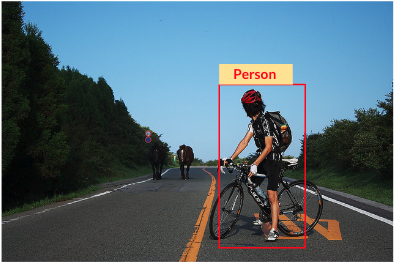}
        \caption{$C^{\mathcal{T}_1}=\{person, \cdots\}$}
        \label{subfig:t1}
    \end{subfigure}
    \begin{subfigure}[t]{0.32\textwidth}
        \centering
        \includegraphics[width=1\linewidth]{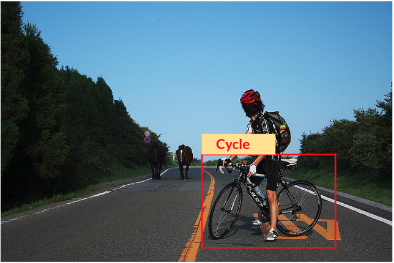}
        \caption{$C^{\mathcal{T}_2}=\{cycle, \cdots\}$}
        \label{subfig:t2}
    \end{subfigure}
    \begin{subfigure}[t]{0.32\textwidth}
        \centering
        \includegraphics[width=1\linewidth]{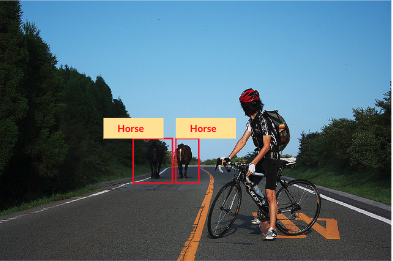}
        \caption{$C^{\mathcal{T}_3}=\{horse, \cdots\}$}
        \label{subfig:t3}
    \end{subfigure}
\vspace{-0.1in}
\caption{{\bf Class-incremental continual object detection.} Each task is characterized by a distinct set of classes, meaning $C^{\mathcal{T}_i} \cap C^{\mathcal{T}_j} = \emptyset$. The category {\em person} in $\mathcal{T}_1$ remains unannotated in all future tasks ($\mathcal{T}_2$ and $\mathcal{T}_3$ in the provided illustration), giving rise to the problem of {\em background relegation} for the object category {\em person}.}
\label{fig:teaser}
\vspace{-0.2in}
\end{figure*}

The majority of techniques for effective continual learning and mitigation of catastrophic forgetting have been developed for continual classification \cite{IL_iCaRL,IL_3,CIL1_zhou2022model,l2p_wang2022learning,smith2023coda,hide_prompt_wang2024hierarchical,ewc_huszar2017quadratic,lwf1_li2017learning}, much fewer approaches have looked at more granular visual tasks, such as object detection \cite{ORE,OW-DETR,cl_detr,erd_feng2022overcoming,cont_iod1} or segmentation \cite{ios_douillard2021plop,ios_gu2021class}.
Modern continual methods designed for such granular tasks, typically integrate a replay buffer with transformer-based architectures \cite{OW-DETR,cl_detr,erd_feng2022overcoming,prob_zohar2023prob} like DETR \cite{detr} and Deformable-DETR \cite{ddetr}, or sometimes the Faster-RCNN backbone \cite{peng2020faster,rosetta_yang2022continual,abr_liu2023augmented,ORE}. Some recent continual object detection methods also incorporate knowledge distillation, sometimes in conjunction with a replay buffer \cite{cl_detr,erd_feng2022overcoming,abr_liu2023augmented,leocod_wu2023label}. 
Despite the successes of these methods, the use of a replay buffer limits their applicability to most practical scenarios. Moreover, a fundamental and largely unaddressed challenge persists with all existing continual detection methods: {\em how to handle the repetition of earlier object categories in future tasks?} For example, consider a scenario where given a continuous stream of data arranged in tasks $\mathcal{T} = \{\mathcal{T}_1, \cdots, \mathcal{T}_t, \cdots, \mathcal{T}_n\}$, the object category {\em person} occurs in $\mathcal{T}_1$ and {\em cycle} occurs in $\mathcal{T}_2$ (as shown in Figure \ref{fig:teaser}). Existing continual detection systems typically remove the annotation for {\em cycle} during training $\mathcal{T}_1$, while {\em person} is removed from $\mathcal{T}_2$. However, removing the {\em person} annotation in $\mathcal{T}_2$ is counterintuitive, as most images of cycles will include a person, causing the most representative class ({\em person}) to be relegated to the background. This becomes a significant issue as transformer-based detection models start treating the {\em person} class as background in future tasks $\{\mathcal{T}_k: k>1\}$. This is a critical problem, which, to our knowledge, has not been addressed.

In this work, we present a transformer network with integrated memory designed to adapt a pre-trained Deformable DETR through continuous fine-tuning. The use of integrated memory has several benefits, including memory and compute efficiency, e.g., as compared to replay buffer models, privacy preservation, and effective catastrophic forgetting mitigation. To enable the latter, we introduce a localized memory retrieval mechanism incorporating a regularization of the query function of memory. This is complemented by gradient masking and background thresholding optimization during training, effectively addressing the background relegation issue. Finally, we assess the proposed method within a continual detection framework established on MS-COCO \cite{mscoco} and PASCAL-VOC \cite{VOC2007} datasets. Through extensive experiments, our results demonstrate that the proposed technique surpasses current state-of-the-art methods by approximately $5-7\%$ on MS-COCO and PASCAL-VOC datasets, establishing a new benchmark for state-of-the-art performance in continual object detection.

\vspace{0.1in}
\noindent
{\bf Contributions:} We propose a novel \textbf{M}emory-augmented \textbf{D}eformable \textbf{DE}tection \textbf{TR}ansformer (\textbf{MD-DETR}) for continual object detection. In doing so, we introduce an integrated memory designed with a flexible local query retrieval mechanism for continual detection. Additionally, we introduce continual optimization strategies to address background relegation issues inherent in continual detection scenarios. The performance of the proposed technique is evaluated on MS-COCO and PASCAL-VOC datasets, demonstrating superior results compared to existing continual detection systems and establishing a new state-of-the-art.

\section{Related Work}

\textbf{Traditional continual learning methods}. 
Traditional methods for continual learning can be broadly classified into three categories: replay buffer \cite{IL_iCaRL,replay1_yoon2021online_coreset,replay2_tiwari2022gcr,replay3_replay_wang2022cafe,rolnick2019experience_replay}, regularization \cite{ewc_huszar2017quadratic,mcdonnell2024ranpac}, and parameter isolation \cite{iso_houlsby2019parameter,mallya2018packnet,iso_p_zhu2023self}. Replay buffer techniques, as mentioned earlier, depend on a sample set (referred to as exemplars) drawn from previous tasks. The replay of this buffer during subsequent tasks is employed to prevent catastrophic forgetting \cite{IL_iCaRL,replay1_yoon2021online_coreset,replay2_tiwari2022gcr,replay3_replay_wang2022cafe,rolnick2019experience_replay}. In contrast, regularization methods impose constraints on gradients to slow down learning on specific weights, considering their importance relative to previous tasks \cite{ewc_huszar2017quadratic}. Knowledge distillation, a more recent form of regularization, applies constraints on the parameter space of the model to ensure that parameters remain in close proximity and do not drift too far across tasks \cite{cl_detr,erd_feng2022overcoming}. On the other hand, parameter isolation methods aim to allocate a subset of the parameters of the machine learning model for each task, optimizing for the efficient allocation of parameters for continual tasks \cite{iso_houlsby2019parameter,mallya2018packnet}.

\vspace{0.15cm}
\noindent \textbf{Memory networks for continual learning}. 
Recently, memory networks have emerged as a solution to address the issue of catastrophic forgetting in pre-trained continual models \cite{l2p_wang2022learning,wang2022dualprompt,smith2023coda,hide_prompt_wang2024hierarchical,DAP_jung2023generating}. The memory is organized as a set of prompts, which are learnable parameters constituting a memory pool. This memory is accessed through an instance-wise query mechanism, with the majority of the pre-trained network remaining frozen to prevent information drift across tasks. L2P \cite{l2p_wang2022learning} and DualPrompt \cite{wang2022dualprompt} employ the selection of top-K memory units for any given inputs based on the cosine similarity of the instance query and the key-value pairs of the memory units. CodaPrompt \cite{smith2023coda} replaces this selection criterion with a more natural mechanism involving a learnable linear combination of memory units. More recently, HiDe-Prompt \cite{hide_prompt_wang2024hierarchical} introduced more general criteria for the hierarchical decomposition of prompts, considering within-task prediction, task-identity inference, and task-adaptive prediction. 
Most memory-based continual solutions only work for image classification, and their applicability to complex vision tasks is still not well established.

\vspace{0.15cm}
\noindent \textbf{Continual/incremental object detection}. 
Addressing continual learning challenges in complex computer vision tasks like detection and segmentation is more intricate compared to classification due to the recurrence of object categories across tasks. Existing solutions heavily depend on replay buffers \cite{ORE,OW-DETR,erd_feng2022overcoming,abr_liu2023augmented,leocod_wu2023label,cl_detr}, with some approaches incorporating knowledge distillation to mitigate forgetting \cite{sid_peng2021sid,erd_feng2022overcoming,cl_detr,abr_liu2023augmented,leocod_wu2023label}. Notably, the recently introduced CL-DETR \cite{cl_detr} integrates a replay buffer with knowledge distillation, demonstrating superior performance compared to earlier distillation-based methods \cite{erd_feng2022overcoming,sid_peng2021sid,ifd_kang2023alleviating}. Researchers have explored various object detection architectures as backbones for continual object detection. Some current solutions still use Faster-RCNN \cite{girshick2015fast} as the continual backbone detector \cite{ORE,rosetta_yang2022continual,abr_liu2023augmented}, many of which rely on a replay buffer to achieve state-of-the-art performance. More recent continual object detectors leverage transformable architectures such as DETR \cite{detr} and Deformable-DETR \cite{ddetr}. For example, OW-DETR \cite{OW-DETR}, PROB \cite{prob_zohar2023prob}, and CL-DETR \cite{cl_detr} utilize Deformable DETR for incremental object detection (along a replay buffer).

\vspace{0.15cm}
\noindent Replay-buffer-based approaches struggle to achieve an optimal balance between the stability and plasticity trade-off, primarily due to the disjoint fine-tuning on the buffer from past tasks, resulting in suboptimal performance on the current task. In contrast, the proposed \textbf{MD-DETR} stands out as a replay-buffer-free method with an improved stability-plasticity trade-off for continual detection when compared to existing state-of-the-art methods such as OW-DETR \cite{OW-DETR}, PROB \cite{prob_zohar2023prob}, and CL-DETR \cite{cl_detr}. Furthermore, the incorporation of memories in a transformer-based continual approach does not effectively address key challenges in complex vision tasks like detection, such as the background relegation of past object categories. Existing memory-based solutions for continual learning, such as L2P \cite{l2p_wang2022learning}, CodaPrompt \cite{smith2023coda}, and DualPrompt \cite{wang2022dualprompt}, are primarily designed for classification and are not well-suited in their current form for the intricacies of complex vision tasks, especially continual detection. Proper utilization of memory networks for continual detection necessitates the implementation of a suitable query retrieval function and continual optimization.

\section{Problem Formulation and Preliminaries}
\noindent \textbf{Continual/incremental object detection}.  In a continual setup, the tasks are perceived in sequential order as they arrive $\mathcal{T} = \{\mathcal{T}_1, \cdots, \mathcal{T}_t, \cdots, \mathcal{T}_n\}$. At a given time step $t$, the dataset is given by $D^{\mathcal{T}_t} = \{X^{\mathcal{T}_t}, Y^{\mathcal{T}_t}\}$, where $X^{\mathcal{T}_t}$ and $Y^{\mathcal{T}_t}$ are images and corresponding annotations. The goal is to train a machine learning model with parameters $\mathcal{\theta}^{\mathcal{T}_t}$ such that the model retains information of all previous tasks $\{1, \cdots, t-1\}$. Note that, $\mathcal{\theta}^{\mathcal{T}_t}$ is initialized with $\mathcal{\theta}^{\mathcal{T}_{t-1}}$, and is fine-tuned on the current task ${\mathcal{T}_t}$. At the end of the training, we obtain a model $\mathcal{\theta}^{\mathcal{T}_n}$, which is used for inference on all tasks.  We follow the class-incremental setup where the object categories at time $t$ are given by $C^{\mathcal{T}_t} \in C$ (where $|C|$ denotes the total number of object categories across all tasks), and incrementally added for each task ${\mathcal{T}_t}$. In other words, the classes are disjoint among tasks, {\em i.e.}, $\forall_{s,t \in C}C^{\mathcal{T}_t} \cap C^{\mathcal{T}_{s \neq t}} = \emptyset$.

\vspace{0.15cm}
\noindent \textbf{Revisiting Deformable-DETR}. The Deformable-DETR \cite{ddetr} introduces deformable attention modules that attend to a small set of key sampling points around a reference. Given an input image $x \in \mathbb{R}^{3 \times H \times W}$, the model $\mathcal{\theta}^{\mathcal{T}_t}$ computes content feature $z_q$ and reference point $p_q$ indexed at multi-head attention query $q$. The deformable attention is given by
\begin{align}
    DA(z_q,p_q,x) = \sum_{m=1}^M \mathbf{W}_m \big[ \sum_{k=1}^K A_{mqk} \cdot \mathbf{W}'_m x(p_q + \delta p_{mqk}) \big]
\end{align}
\noindent where $m$ indexes the attention head, $k$ indexes the sampled keys, and $K$ is the total sampled key number. Model $\mathcal{\theta}^{\mathcal{T}_t}$ predicts object categories $\{\hat{y}_j; j \in C^{\mathcal{T}_t}\}$ using the deformable attention and self-attention layers. The prediction $\hat{y}_j = \{\hat{s}_j, \hat{b}_j\}$ consists of the confidence score ($\hat{s}_j$) or class probability for the predicted object category $j$, while $\hat{b}_j$ is the bounding box predictions. To handle object proposals without ground truth or background information, an extra logit is appended to the score vector, resulting in $|\hat{s}_j| = |C^{\mathcal{T}_t}|+1$. To adapt the background logit to a continual setup, the same logit is assigned as the background bit for all tasks, ensuring the cardinality of the classification head becomes $|C|$+1.

Model $\theta^{\mathcal{T}_t}$ is trained end-to-end optimizing the loss:
\begin{align}
    \mathcal{L}_{detr} = \sum_i -\log(\hat{p}_{\sigma_i}, s_i) + \mathbb{I}_{c(s_i) \neq \phi} \mathcal{L}_{box} (\hat{b}_{\sigma_i}, b_i)
\end{align}
where, $\mathcal{L}_{box} (\hat{b}_{\sigma_i}, b_i)$ is the bounding box prediction and $\hat{\sigma}$ is the optimal assignment of ground truth labels to the object predictions $\mathcal{H}$, which is obtained by solving the set prediction criterion using the Hungarian loss \cite{kuhn1955hungarian}.
\begin{align}
    \hat{\sigma} = \mathcal{H}(y_i, \hat{y}) = \underset{\sigma\in S_N}{argmax} \sum_{i\in \mathcal{N}} \mathbb{I}_{c(s_i) \neq \phi} \Big\{ -\log(\hat{p}_{\sigma_i}, s_i)| + \mathcal{L}_{box} (\hat{b}_{\sigma_i}, b_{\sigma_i}) \Big\}
\end{align}
where $y_i \in \{s_i,b_i\}$ is the ground truth and $\hat{y} \in \{\hat{p}_{\sigma_i},\hat{b}_i\}$ is the prediction from $\mathcal{\theta}^{\mathcal{T}_t}$.

\noindent \textbf{Prompt-based memory networks}. Prompt-tuning or prefix-tuning-based memory augmented networks utilize a shared pool of memory units, often referred to as a prompt-pool in existing literature \cite{l2p_wang2022learning,wang2022dualprompt,smith2023coda,hide_prompt_wang2024hierarchical,DAP_jung2023generating}, to address catastrophic forgetting in continual learning. The fundamental concept involves employing a key-value pair-based query retrieval system that dynamically selects relevant memory units for a given input sample, with each memory unit associated with a learnable key. The memory retrieval process involves mapping the query function to the key through a similarity function, typically cosine similarity. The retrieved memory units are usually integrated within the multi-head attention layer in the decoder layers. The learning of memory modules can be achieved through a similarity-based regularization criterion, as demonstrated in L2P \cite{l2p_wang2022learning}, or a more general weighting scheme, as introduced by \cite{smith2023coda}. In this study, we define memory modules like L2P \cite{l2p_wang2022learning} and CodaPrompt \cite{smith2023coda}; however, our emphasis is on addressing challenges specific to complex computer vision tasks such as detection.
\begin{figure*}[!t]
        \centering
        \includegraphics[scale=0.18]{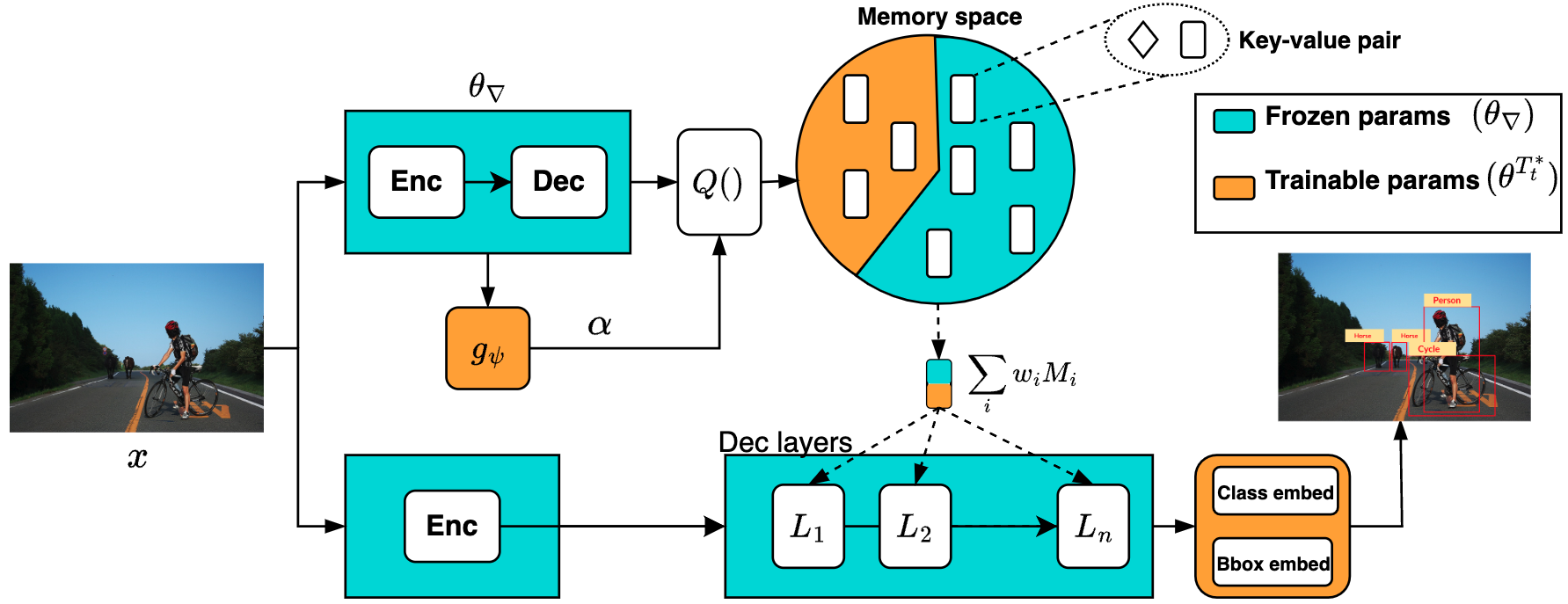}
        \caption{{\bf Architecture of MD-DETR at a given time-step $t$.} Given an input image $x$, we use proposed query function $Q(x,\theta_\nabla,\alpha)$ to retrieve relevant memory units as a linear combination. The obtained information from the memory is utilized by the decoder across various decoding layers. The majority of the architecture remains frozen, encompassing the encoder and decoder; the trainable modules consist of memory units $\mathbf{M}$, class embedding, bounding box embedding, and ranking function $g_\psi$.}
        \label{fig:model}
        \vspace{-4pt}
\end{figure*}
\section{Proposed Method}
We design our architecture by freezing the encoder and decoder layers of a pre-trained Deformable-DETR (to be denoted by parameters $\theta_\nabla$). The learnable parameters (to be denoted by $\theta^{{\mathcal{T}_t}^*}$) are the class embedding, bounding box embedding, and the memory modules which we will discuss next. The parameter space of the deformable-DETR architecture is defined by: $\theta^{\mathcal{T}_t} = \theta_\nabla \cup \theta^{{\mathcal{T}_t}^*}$. (Note that for the remainder of this section, we will employ $\theta$ to refer to the model itself. This choice is made to minimize the introduction of additional notation.)

\subsection{Memory modules for Deformable-DETR}
Let $\mathbf{M} \in \mathbb{R}^{L_m \times D \times N_m}$ be the shared pool of memory units with the number of units $N_m$, memory length $L_m$, and dimension $D$. Each memory unit is associated with a learnable keys $\mathbf{K}$: $\{(\mathbf{k}_1,\mathbf{M}_1), \cdots , (\mathbf{k}_{N_m},\mathbf{M}_{N_m})\}$, where $\mathbf{k}_i \in \mathbb{R}^{D}$. 
The read operation from the memory is defined by a query retrieval function $Q(x, \phi): \mathbb{R}^{3 \times H \times W} \rightarrow \mathbb{R}^{1 \times D}$, where $x\in \mathbb{R}^{3 \times H \times W}$ is given input image, and $\phi$ is a deterministic feature extractor. We leverage the frozen layers of our pre-trained network to create the query function $Q(x,{\theta_\nabla})$ for retrieving memory units: $\rho(Q(x,{\theta_\nabla}), k_m)$, where $\rho$ denotes the similarity function. In our approach, we incorporate learnable parameters $\mathbf{A} \in \mathbb{R}^{D \times N_m}$ that generate a weight vector $w$ to perform the memory read operation ($\mathbf{M}^{r}$):
\begin{align}
    w = \rho (Q(x,{\theta_\nabla})\odot A,\mathbf{K}) &= \{\rho(Q(x,{\theta_\nabla})\odot A_1, \mathbf{k}_1), \cdots , \rho(Q(x,{\theta_\nabla})\odot A_{N_m}, \mathbf{k}_1)\} \\
    \mathbf{M}^{r} &= \sum_i w_i \mathbf{M}_i
\end{align}
where $\mathbf{M}^{r}$ represents a linear combination of all memory units weighted by $w$ and conditioned on $Q(x,{\theta_\nabla})$. The incorporated information retrieved from the memory is integrated into the deformable attention module of the decoder layer as follows:
\begin{align}
    DA_{M}(z_q,p_q,x, \textbf{M}^{r}) &= \sum_{m=1}^M \mathbf{W}_m \big[ \sum_{k=1}^K A_{mqk} (\mathbf{M}^r) \cdot [\mathbf{W}'_m:\mathbf{M^r}{[\frac{L_m}{2}:,,]}] x(p_q + \delta p_{mqk}) \big] \\
    A_{mqk} (\mathbf{M}^r) &\propto  \exp { \frac{z_q^U U^T_m [V_m:\mathbf{M^r}{[:\frac{L_m}{2},,]}] x_k}{\sqrt{C_v}} }
\end{align}
\noindent that is, we concatenate ($[x:y]$) the first half of retrieved memory $\mathbf{M^r}{[:\frac{L_m}{2},,]}$ to the input key projections matrix at $m^{th}$ head ($\mathbf{W}'_m$) and $\mathbf{M^r}{[\frac{L_m}{2}:,,]}$ to the value projections at the $m^{th}$ head $(V_m)$. 

\subsection{Query function for localized memory retrieval}

Given an input image $x$ and ground truth bounding boxes $b$, the deformable-DETR model produces a set of proposals $\mathbf{P} \in \mathbb{R}^{P \times D}$ in a single pass through the decoder ($P$ >{}> $b$). A basic query function could be constructed by averaging all proposals, expressed as $Q(x,{\theta_\nabla}): x \in \mathbb{R}^{C \times H \times W} \rightarrow \theta_\nabla(x) \in \mathbb{R}^{P \times D}\rightarrow Q(x,{\theta_\nabla}) \in \mathbb{R}^{ 1 \times D}$. However, this approach fails to effectively localize relevant proposals for the given input sample and may be ineffective in capturing useful instance-specific information. A more effective alternative involves selecting a small set of relevant proposals before averaging them. Let $g_\psi$ be a learnable function parameterized by $\psi$ which learns to rank proposals ($\alpha$) based on its relevance given the input $x$: $\alpha = g_\psi(\theta_\nabla(x))\in \mathbb{R}^{1\times P}$, where $\alpha_i$ is the weight given to proposal $P_i \in \mathbb{R}^{1\times P}$. Then, the localized query retrieval can be defined as a function of $\alpha$: $Q(x,\theta_\nabla,\alpha)$. To optimize $\alpha$, we form a cross-entropy regularization using the optimal assignment ($\hat{\sigma}$) obtained from the Hungarian matching criterion as the ground truth. We define the localized query retrieval ($Q(x,\theta_\nabla,\alpha)$) and loss ($\mathcal{L}_Q$) as:
\begin{align}
    Q(x,\theta_\nabla,\alpha) = \sum_i \alpha_i * \{\theta_\nabla(x)\}_i ;\ \mathcal{L}_Q = l(H(\hat{y},y),g_\psi(x)) = CE(\hat{\sigma},\alpha)
\end{align}
\noindent where $\{\theta_\nabla(x)\}_i$ is the $i^{th}$ proposal; $y \in \{s,b\}$ is the ground truth and $\hat{y} \in \{\hat{p}_{\sigma},\hat{b}\}$ is the prediction from $\mathcal{\theta}^{\mathcal{T}_t}$. 

\subsection{Continual optimization for $\theta^{{\mathcal{T}_t}^*}$}
Learnable parameters $\theta^{{\mathcal{T}_t}^*}$ consist of three primary components: memory network, class embedding, and bounding-box embedding. Each of these requires meticulous optimization to operate efficiently within a continuous setting.

\noindent \textbf{Memory network}. Catastrophic forgetting within the memory network occurs when information in $\mathbf{M} \in \mathbb{R}^{L_m \times D \times N_m}$ is overwritten during backpropagation. To mitigate forgetting, we partition the memory units into task-specific chunks, where each chunk's size is defined as: $J^{\mathcal{T}_t} = \frac{N_m}{|\mathcal{T}|}$. We then freeze the memory chunks not associated with the current task $\mathcal{T}^t$, ensuring that only the memory chunk relevant to the current task stores task-specific information: $\mathbf{M}{[:,:, J^{\mathcal{T}{t-1}}:J^{\mathcal{T}_t}]} \in \theta^{{\mathcal{T}_t}^*}$.

\noindent \textbf{Class embedding}. In the context of a class-incremental setup, at any given time-step $t$, the class embedding (class logits of the classification head) can be partitioned into two groups: logits for the previous tasks ($C^{\mathcal{T}_1}, \cdots, C^{\mathcal{T}_t-1}$), and logits for the future tasks ($C^{\mathcal{T}_t+1}, \cdots, C^{\mathcal{T}n}$). For logits corresponding to future tasks, we set the model-predicted score to negative infinity: $\forall_{j\in t+1}^{t_n} s_j = -\infty$. This leads to a softmax value of $0$ for any future class, ensuring they do not impact the optimization of the current task. The same operation can be applied to past classes, but a more effective approach is to mask the gradient for the weights of past classes, preventing the class embedding of previous classes from being updated.

\noindent \textbf{Bounding-box embedding}. The bounding box embedding for a pre-trained model is tuned to different object categories, considering their sizes and shapes. Typically, a model pre-trained on a large dataset has encountered numerous potential object shapes and sizes relevant to a downstream task. Consequently, frequent updates to the bounding box embeddings are unnecessary. We employ a lower learning rate for the bounding box embeddings compared to other learnable parameters. In our experiments, we decrease the learning rate by a factor of 10 for parameters related to bounding boxes.

\subsection{Continual optimization for solving background relegation}
As previously noted, background relegation occurs in a continual detection setup when objects from the most representative class are pushed to the background. While replay-based continual detection systems offer some defense against background relegation, these systems struggle to address unannotated past object categories during training on the current task. Furthermore, the issue persists irrespective of whether a small or large buffer is employed, with a larger buffer leading to potential information loss on the current task. Conversely, non-replay-buffer-based continual methods are vulnerable to background relegation.

We introduce a straightforward heuristic to address this issue. At a given time-step $t$ where $t>1$, we utilize the model $\theta^{\mathcal{T}t}$ to perform inference on each sample in the training batches, resulting in the prediction: $\hat{y_j} = \{\hat{s_j},\hat{b}_j\}$. The category label is determined by the $argmax$ of scores across all categories: $l = argmax_{j \in C}(\hat{s}_j)$. Using the $\hat{y}_j$, we employ two criteria to generate pseudo-labels:
\begin{align}
    l \in \{C^{\mathcal{T}_1}, \cdots, C^{\mathcal{T}_{t-1}}\}; \hat{s}_j > \delta_{bt}
\end{align}

\noindent where $\delta_{bt}$ represents a user-defined threshold. The first criterion ensures that only those unannotated objects are retained, which the model has encountered in the past, while the second criterion filters out predictions where the model is under-confident, thereby reducing the optimization risk. For each batch, we incorporate the pseudo-labels into the training batches, which are then optimized concurrently. We call this procedure {\em background thresholding} (to be denoted by $BT$). We empirically study the effect of $\delta_{bt}$ in the experimentation section.

\subsection{Training loss}
Our training objective consists of $\mathcal{L}_{detr}$ loss and localized query retrieval regularization $\mathcal{L}_Q$:
\begin{align}
    \mathcal{L} = \mathcal{L}_{detr} + \lambda_{Q} \mathcal{L}_{Q}
\end{align}
\noindent where $\lambda_Q$ is the weight given to $\mathcal{L}_{Q}$.

\section{Experiments}

\noindent \textbf{Datasets}. We evaluate our proposed method on MS-COCO \cite{mscoco} and PASCAL-VOC \cite{VOC2007}. We follow the setup given by \cite{OW-DETR} and divide MS-COCO into 4 tasks based on semantic splits, with each task containing a distinct set of classes (refer to supplementary material for further details). This continual setup adopts a multi-step approach, where 80 classes are divided into four sequential streams. Similarly, for PASCAL-VOC, we adhere to the configuration utilized by \cite{prob_zohar2023prob}, evaluating the proposed model in a two-step design consisting of 2 tasks: A+B; where A represents the number of classes introduced in $\mathcal{T}_1$, and B represents the number of classes introduced in $\mathcal{T}_2$. The evaluation is conducted in three settings: 10+10, 15+5, and 19+1. 
The training data is constructed from the trainset, while the evaluation is performed on the val-set for MS-COCO, test-set for PASCAL-VOC, following the evaluation protocol used by \cite{OW-DETR}.

\vspace{0.15cm}
\noindent \textbf{Continual metrics}. We use the standard mean average precision at 50\% intersection over union (mAP@IoU=0.5) as the base metric. Using mAP@IoU=0.5 we define three metrics for continual detection based on the classes introduced for the evaluation:
\begin{align}
    mAP@P \in \{C^{\mathcal{T}_1} \cdots  C^{\mathcal{T}_{t-1}}\}; mAP@C \in \{C^{\mathcal{T}_t}\}; mAP@A \in \{C^{\mathcal{T}_{1}} \cdots  C^{\mathcal{T}_{t}}\}
\end{align}
\noindent where, $P$ = {\em previously seen classes}, $C$ = {\em classes introduced in the current task}, and $A$ = {\em all seen classes till now, i.e, P+C}. The assessment of forgetting at task $\mathcal{T}_t$ is conducted through $mAP@P$, measuring the extent to which information about previously encountered classes is lost during the fine-tuning of the model on $C^{\mathcal{T}_t}$. Similarly, $mAP@C$ and $mAP@A$ help in evaluating the stability-plasticity trade-off. 

\vspace{0.15cm}
\noindent \textbf{Implementation Details}. \textbf{MD-DETR} is built on top of Deformable-DETR \cite{ddetr}, utilizing the same transformer hyper-parameters as used in \cite{ddetr}. We use the Deformable-DETR implementation provided in the HuggingFace repository \cite{wolf2019huggingface}. For all our experiments, we use 100 memory units: $N_m=100$; the length of memory units is 10: $L_m=10$; the dimension of memory units is 256 (which is also the dimension of embedding used in \cite{ddetr}): $D=256$. We use $\lambda_Q$=0.01, and $\delta_{bt}=0.65$ after empirical analysis on a hold-out dataset. (Please refer to the supplementary material for details.)

\subsection{Results}
\textbf{Evaluating performance on MS-COCO}. We present the result on multi-step class incremental setting on MS-COCO in Table \ref{tab:sota_mscoco}. All the methods under comparison (ORE-EBUI \cite{ORE}, OW-DETR \cite{OW-DETR}, CL-DETR \cite{cl_detr}, PROB \cite{prob_zohar2023prob}, ERD \cite{erd_feng2022overcoming}) employ a replay buffer to counteract catastrophic forgetting. In contrast, \textbf{MD-DETR} stands out as the sole buffer-free method. Across all four tasks, \textbf{MD-DETR }preserves a substantial amount of information from previous classes, as indicated by the highest $mAP@P$ value. Noteworthy is \textbf{MD-DETR}'s significant outperformance of PROB and CL-DETR by approximately 10\% on Task 4, highlighting our effectiveness in mitigating catastrophic forgetting.

Continual learning methods face a trade-off between stability (preserving past information) and plasticity (generalizing to current information). Employing a replay buffer poses a drawback of disjoint fine-tuning of past buffers, leading to a decline in performance on the current task, as replaying samples from previous tasks introduces a bias toward those tasks. As demonstrated in Table \ref{tab:sota_mscoco}, strategies utilizing replay buffers struggle to maintain an effective balance between plasticity and stability. In contrast, high $mAP@C$ and $mAP@A$ values for \textbf{MD-DETR} show a balanced stability-plasticity tradeoff, attributed to its architectural and optimization design.

CL-DETR \cite{cl_detr} and ERD \cite{erd_feng2022overcoming} present their results in a multi-step setting on MS-COCO with a 4-step increment, that is similar to our setup. However, their assessment is less robust, focusing solely on reporting $mAP@A$, which combines performance on both current and previous classes. As illustrated in Table \ref{tab:sota_mscoco}, a comprehensive evaluation of continual object detection should incorporate a separate metric for measuring the forgetting of previous classes.

\begin{table*}[t]
\centering
\setlength{\tabcolsep}{3pt}
\adjustbox{width=\textwidth}{
\begin{tabular}{@{}l|c|ccc|ccc|ccc@{}}
\toprule
 \textbf{Task IDs} ($\rightarrow$)& \multicolumn{1}{c|}{\textbf{Task 1}} & \multicolumn{3}{c|}{\textbf{Task 2}} & \multicolumn{3}{c|}{\textbf{Task 3}} & \multicolumn{3}{c}{\textbf{Task 4}} \\ \midrule
 

& \begin{tabular}[c]{c}mAP@C\end{tabular} & \begin{tabular}[c]{@{}c}mAP@P\end{tabular} & \begin{tabular}[c]{@{}c}mAP@C\end{tabular} & mAP@A & \begin{tabular}[c]{@{}c@{}}mAP@P\end{tabular} & \begin{tabular}[c]{@{}c@{}}mAP@C\end{tabular} & mAP@A & \begin{tabular}[c]{@{}c@{}}mAP@P\end{tabular} & \begin{tabular}[c]{@{}c@{}}mAP@C\end{tabular} & mAP@A \\ \midrule

ORE $-$ EBUI \cite{ORE} & 61.4 & 56.5 & 26.1 & 40.6  & 37.8 & 23.7 & 33.7 & 33.6 & 26.3 & 31.8 \\ 
OW-DETR \cite{OW-DETR} & 71.5 & 62.8 & 27.5 & 43.8  & 45.2 & 24.4 & 38.5 & 38.2 & 28.1 & 33.1 \\
PROB \cite{prob_zohar2023prob} & 73.4 & 66.3 & 36.0 & 50.4  & 47.8 & 30.4 & 42.0 & 42.6 & 31.7 & 39.9 \\ 
CL-DETR \cite{cl_detr} & - &- & - & -  & - & - & - & - & - & 39.2 \\ 
ERD \cite{erd_feng2022overcoming} & - &- & - & -  & - & - & - & - & - & 35.4 \\ 

\midrule
\midrule 
\textbf{MD-DETR} & \textbf{78.5} &\textbf{69.1} & \textbf{56.5} &\textbf{61.2} & \textbf{54.6} & \textbf{58.3} & \textbf{55.4} &\textbf{51.5} & \textbf{52.7} & \textbf{50.2} \\ 
\bottomrule

\end{tabular}%
}
\caption{\textbf{Evaluating the state-of-the-art performance for continual detection on MS-COCO.}  The results are reported in a multi-step setting. \vspace{-0.3cm}}
\label{tab:sota_mscoco}
\end{table*}

\begin{table*}[t]
\centering
\setlength{\tabcolsep}{3pt}
\adjustbox{width=\textwidth}{
\begin{tabular}{@{}l|ccc|ccc|ccc@{}}
\toprule
 \textbf{A+B setting} ($\rightarrow$) & \multicolumn{3}{c|}{\textbf{10+10}} & \multicolumn{3}{c|}{\textbf{15+5}} & \multicolumn{3}{c}{\textbf{19+1}} \\ \midrule
 

& \begin{tabular}[c]{@{}c}mAP@P\end{tabular} & \begin{tabular}[c]{@{}c}mAP@C\end{tabular} & mAP@A & \begin{tabular}[c]{@{}c@{}}mAP@P\end{tabular} & \begin{tabular}[c]{@{}c@{}}mAP@C\end{tabular} & mAP@A & \begin{tabular}[c]{@{}c@{}}mAP@P\end{tabular} & \begin{tabular}[c]{@{}c@{}}mAP@C\end{tabular} & mAP@A \\ \midrule

ILOD \cite{ILOD} & 63.2 & 63.2 & 63.2 & 68.3  & 58.4 & 65.8 & 65.8 & 62.7 & 68.2 \\ 
Faster ILOD \cite{peng2020faster} & 69.8 & 54.5 & 62.1  & 71.6 & 56.9 & 67.9 & 68.9 & 61.1 & 68.5 \\ 
ORE $-$ EBUI \cite{ORE} & 60.4 & 68.8 & 64.5  & 71.8 & 58.7 & 68.5 & 69.4 & 60.1 & 68.8 \\ 
OW-DETR \cite{OW-DETR} & 63.5 & 67.9 & 65.7 & 72.2 & 59.8 & 69.4 & 70.2 & 62.0 & 70.2 \\ 
PROB \cite{prob_zohar2023prob} & 66.0 & 67.2 & 66.5  & 73.2 & 60.8 & 70.1 & 73.9 & 48.5 & 72.6 \\ 
ROSETTA \cite{rosetta_yang2022continual} & - & - & 66.8  & - & - & 69.1 & - & - & 69.6 \\ 

\midrule
\midrule 
\textbf{MD-DETR} & \textbf{73.1} & \textbf{77.5} & \textbf{73.2} & \textbf{77.4}  & \textbf{69.4} &\textbf{ 76.7} & \textbf{76.8} & \textbf{67.2} & \textbf{76.1} \\ 
\bottomrule

\end{tabular}%
}
\caption{\textbf{State-of-the-art comparison for continual detection on PASCAL-VOC.} We report the results in the in the two-step A+B setting. \vspace{-0.3cm}}
\label{tab:sota_voc}
\end{table*}

\vspace{0.15cm}
\noindent \textbf{Evaluating performance on PASCAL-VOC}. 
We present the outcomes in a two-step class incremental scenario using PASCAL-VOC \cite{VOC2007}, which is a less stringent setup compared to the multi-step stream of MS-COCO. In each evaluation, the model is first trained on 10/15/19 object classes, and then an additional 10/5/1 classes are incrementally introduced. As shown in Table \ref{tab:sota_voc}, \textbf{MD-DETR} outperforms all state-of-the-art methods by a large margin. Among the compared methods, ROSETTA \cite{rosetta_yang2022continual} is the sole continual detection method that employs a replay-free strategy.

\begin{table*}[t]
\centering
\setlength{\tabcolsep}{3pt}
\adjustbox{width=\textwidth}{
\begin{tabular}{@{}l|c|ccc|ccc|ccc@{}}
\toprule
 \textbf{Task IDs} ($\rightarrow$)& \multicolumn{1}{c|}{\textbf{Task 1}} & \multicolumn{3}{c|}{\textbf{Task 2}} & \multicolumn{3}{c|}{\textbf{Task 3}} & \multicolumn{3}{c}{\textbf{Task 4}} \\ \midrule
 

& \begin{tabular}[c]{c}mAP@C\end{tabular} & \begin{tabular}[c]{@{}c}mAP@P\end{tabular} & \begin{tabular}[c]{@{}c}mAP@C\end{tabular} & mAP@A & \begin{tabular}[c]{@{}c@{}}mAP@P\end{tabular} & \begin{tabular}[c]{@{}c@{}}mAP@C\end{tabular} & mAP@A & \begin{tabular}[c]{@{}c@{}}mAP@P\end{tabular} & \begin{tabular}[c]{@{}c@{}}mAP@C\end{tabular} & mAP@A \\ \midrule

FT & 78.1 & 59.2 & 54.2 & 40.3  & 43.7 & 55.4 & 46.7 & 21.6 & 49.5 & 23.7 \\ 
FT+Mem & 78.5 & 63.5 & 55.8 & 56.4  &45.3 &57.4 & 48.4 & 28.2 & 50.7 & 30.3 \\ 
FT+Mem+BT & 78.5 & 66.2 & 55.8 & 57.6  & 48.4 & 57.7 & 50.1 & 35.2 & 50.9 & 38.8 \\ 
FT+Mem+QL & 78.5 & 67.2 & 56.1 & 59.5  & 49.2 & 58.2 & 50.5 & 39.6 & 51.9 & 42.3 \\ 
FT+Mem+BT+QL & \textbf{78.5} &\textbf{69.1} & \textbf{56.5} &\textbf{61.2} & \textbf{54.6} & \textbf{58.3} & \textbf{55.4} &\textbf{51.5} & \textbf{52.7} & 
\textbf{50.2} \\ 
\midrule
\midrule

MD-DETR ($\delta_{bt}=0.25$)& - & 65.3 & 54.5 & 56.2  & 50.1 & 56.3 & 50.7 & 42.6 & 50.5 & 41.5 \\ 
MD-DETR ($\delta_{bt}=0.50$) & - & 68.7 & 56.5 & 61.0  & 54.2 & 58.2 & 55.1 & 51.4 & 52.7 & 50.1 \\ 
MD-DETR ($\delta_{bt}=0.65$) & -& \textbf{69.1} & \textbf{56.5} &\textbf{61.2} & \textbf{54.6} & \textbf{58.3} & \textbf{55.4} &\textbf{51.5} & \textbf{52.7} & 
\textbf{50.2} \\ 
MD-DETR ($\delta_{bt}=0.85 $) & -& 68.2 & 56.4 & 60.7  & 51.5 & 58.1 & 53.4 & 49.4 & 52.5 & 48.3 \\ 

\bottomrule

\end{tabular}%
}
\caption{\textbf{Ablation study on various components of MD-DETR on MS-COCO.} The top block shows the effect of adding different components to the fine-tuning backbone ($FT$): memory units ($Mem$), background thresholding ($BT$), and localized query retrieval ($QL$). The bottom block shows the effect of various background thresholding values ($\delta_{bt}$) on $FT+Mem+BT+QL$. \vspace{-0.3cm}}
\label{tab:ablate}
\end{table*}

\begin{figure*}[!t]
        \centering
        \includegraphics[scale=0.20]{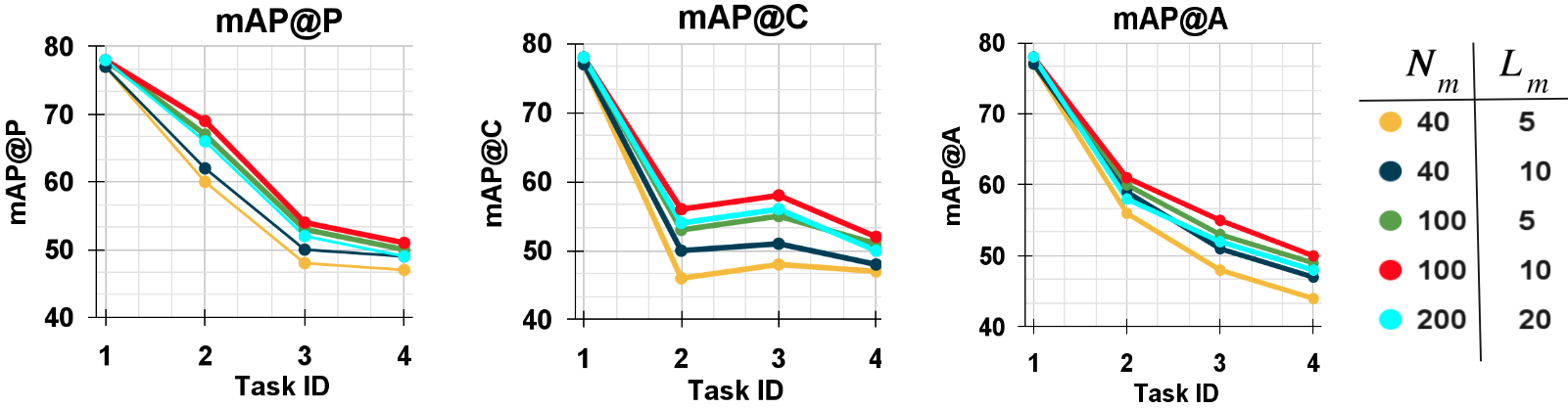}
        \caption{{\bf Studying the effect of varying the number of memory units $N_m$ and length of memory $L_m$.} The ablation is conducted on MS-COCO.}
        \label{fig:ablate_mem}
        \vspace{-4pt}
\end{figure*}

\subsection{Ablations}

We examine the importance of each component in the proposed model, as detailed in Table \ref{tab:ablate}. The $FT$ architecture represents the fine-tuned version of \textbf{MD-DETR} without memory modules and background relegation. Subsequently, we investigate the impact of incorporating memory units ($Mem$), background thresholding ($BT$), and localized query retrieval ($QL$). The introduction of memory units leads to a reduction in forgetting, that is higher $mAP@P$, with more pronounced improvements in subsequent tasks. Similarly, the inclusion of background thresholding improves $mAP@P$, while localized query retrieval enhances both $mAP@P$ and $mAP@C$. The combination of localized query retrieval and background thresholding yields the best performance, surpassing all existing state-of-the-art methods. 
\begin{figure*}[!t]
        \centering
        \includegraphics[scale=0.24]{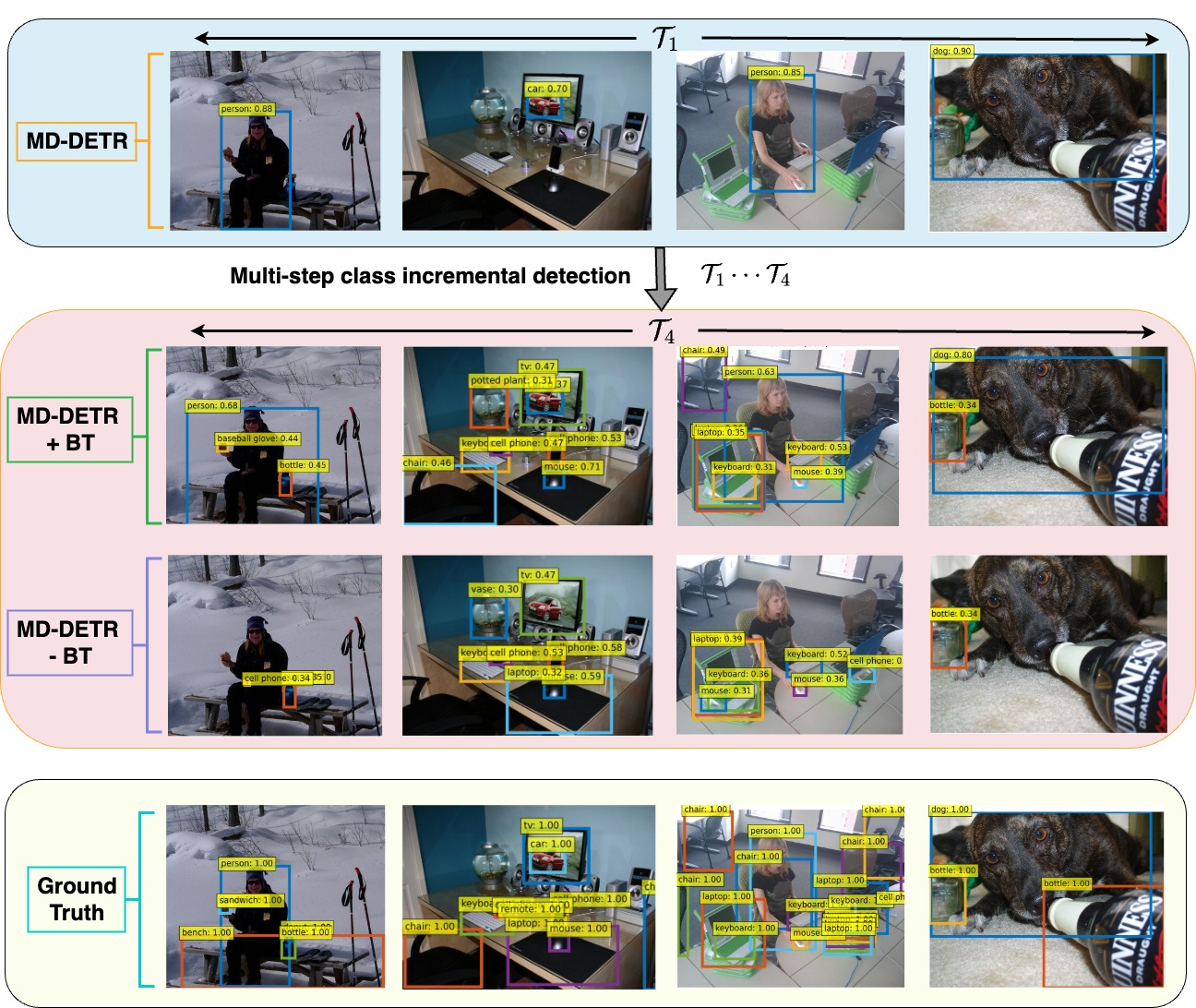}
        \caption{{\bf Progression of MD-DETR performance from $\mathcal{T}_1 \rightarrow \mathcal{T}_4$ when trained in a multi-step class incremental setting on MS-COCO.} To illustrate the effectiveness of MD-DETR in addressing background relegation of previously encountered classes, a comparison is presented between two architecture designs: with (\textbf{MD-DETR} + \textbf{BT}) and without (\textbf{MD-DETR} - \textbf{BT}) background thresholding. In the images shown in the first column (1$^{st}$ image for each block), the class {\em \{person\}} is relegated by \textbf{MD-DETR} - \textbf{BT}; similarly, for the second column, the category {\em \{car\}} is relegated, the third column displays relegation of classes {\em \{person, chair\}}, and the fourth column shows the relegation of the category {\em \{dog\}}. The ground truth block shows images with all annotations across all the four tasks $\{\mathcal{T}_1 \cdots \mathcal{T}_4\}$.
        }
        \label{fig:viz_cont}
        \vspace{-4pt}
\end{figure*}
Subsequently, we examine the impact of background thresholding on \textbf{MD-DETR}, experimenting with different values for $\delta_{bt} \in \{0.25,0.50,0.65,0.85\}$. Lower values of $\delta_{bt}$ (<0.40) increase the likelihood of false positive pseudo-labels, leading to decreased values for both $mAP@P$ and $mAP@C$. Conversely, employing a very high value of $\delta_{bt}$ (>0.80) significantly reduces the chances of sampling pseudo-labels, once again resulting in lower values for $mAP@P$ and $mAP@C$. We identify values approximately around $\delta_{bt} \in \{0.6 \pm 0.1\}$ as suitable candidates for achieving good continual detection performance.

We also explore different hyperparameters employed in the design of our memory network, namely $N_m$ and $L_m$. As shown in Figure \ref{fig:ablate_mem}, 
we assess the performance of different configurations across multiple metrics—$mAP@P$, $mAP@C$, $mAP@A$—across four tasks in a multi-step continual setup on MS-COCO. Decreasing the memory capacity leads to a decline in $mAP@P$, as does reducing the length. Similarly, performance degrades when employing a large memory pool ($N_m=200$) with an extended length $L_m=20$, due to optimization instability. The optimal performance is observed with $N_m=100$ and $L_m=10$.
\subsection{Discussion}

We present a qualitative illustration of \textbf{MD-DETR} on continual object detection on MS-COCO in Figure \ref{fig:viz_cont}. The illustration depicts the evolution of \textbf{MD-DETR}'s effectiveness from task 1 ($\mathcal{T}_1$) to task 4 ($\mathcal{T}_4$). We specifically explore the impact of background relegation and analyze its effects at $\mathcal{T}_4$. The classes encountered in $\mathcal{C}^{\mathcal{T}_1} \in \{person, car, dog, \cdots \}$, while the classes encountered in $\mathcal{C}^{\mathcal{T}_4} \in \{bottle, computer, laptop, tv, car, dog, \cdots \}$. The \textbf{MD-DETR }architecture equipped with background thresholding (\textbf{MD-DETR} + \textbf{BT}) effectively retains instances from the previous tasks. In contrast, without background thresholding (\textbf{MD-DETR} - \textbf{BT}), most classes observed in $\mathcal{T}_1$ are forgotten due to the background relegation of these classes in $\mathcal{T}_4$. For a comprehensive qualitative analysis of individual tasks, please refer to the supplementary section.

\vspace{0.15cm}
\noindent \textbf{Limitations}. 
The proposed approach \textbf{MD-DETR} effectively addresses the challenges in continual detection, surpassing all existing state-of-the-art methods. This success can be attributed primarily to its architectural design and the continual optimization that mitigates background relegation issues. However, we note two other fundamental challenges that remain unattended in continual detection methods, including \textbf{MD-DETR}: bounding-box deformation and confidence reduction on previously encountered classes. Catastrophic forgetting in continual detection systems not only impacts class-level information but also affects the preservation of bounding box shapes. In certain instances, the bounding box configurations for most classes seen in the past may exhibit shapes different from those of the new object categories, resulting in the deformation of bounding boxes related to old object categories. Similarly, we also observe a reduction in confidence scores for previously seen classes during future evaluations. Currently, \textbf{MD-DETR} does not address these two issues, as depicted in Figure \ref{fig:viz_cont}.
Both of these challenges stem from the stability-plasticity dilemma and cannot be effectively addressed by incorporating a replay buffer. A smaller buffer proves inadequate in preserving past bounding box shapes and maintaining high confidence scores. Conversely, a larger buffer tends to substantially shift predictions toward previous categories, resulting in poor performance on current classes.


\section{Conclusion}
In this study, we introduce \textbf{MD-DETR}, a memory-augmented transformer architecture designed for continual detection. This method stands out as a robust non-buffer-based strategy that effectively addresses catastrophic forgetting in continual detection. Additionally, we identify a fundamental issue of background relegation present in all existing continual learning techniques and propose an effective mitigation strategy within our framework. Furthermore, we highlight two consequences of the stability-plasticity dilemma in continual detection scenarios: bounding box deformation and a reduction in confidence scores for previously encountered classes. Addressing these challenges represents a relevant research direction toward enhancing the overall robustness of continual detection systems.

\section*{Acknowledgements}
The computation resources used in preparing this research were provided by the Digital Research Alliance of Canada, and The Vector Institute for AI, Canada.

\clearpage  

\appendix

\begin{center}
    \Large
    \textbf{Preventing Catastrophic Forgetting through Memory Networks in Continuous Detection (Supplementary Material)}
\end{center}


\section{Implementation Details}
Our Deformable-DETR backbone is trained using the original hyperparameters outlined in \cite{ddetr}, excluding their iterative bounding box refinement and the two-stage mechanism. Specifically, we set the number of proposals ($\mathbf{P}$) to 300, with 6 encoder/decoder layers, 8 encoder/decoder attention heads, and a dimension ($\textbf{D}$) of 256. We use the Adam optimizer with a learning rate ($lr$) of 0.01 and weight decay of 0.0004 for all learnable parameters, except for bounding boxes, which utilize a learning rate of 0.001. On MS-COCO, we train \textbf{MD-DETR} for 10 epochs for $\mathcal{T}_1$, while the number of epochs for ${\mathcal{T}_2 \cdots \mathcal{T}_4}$ is 5. On VOC, we train \textbf{MD-DETR} for 45 epochs for both the A and B tasks. 

In our memory network, the value of the number of memory units ($N_m$) is set to 100 and a memory length ($L_m$) to 10, based on the findings from Figure 3 in the main draft. Likewise, the background thresholding value ($\delta_{bt}$) is set to 0.65, determined from our analysis presented in Table 3 of the main draft. We select the value of $\lambda_Q$ to be 0.01.

\vspace{0.1cm}
\noindent \textbf{Pre-Training}. We utilize the HuggingFace transformers repository \footnote{https://huggingface.co/} to construct our framework. Initially, we employ the Deformable-DETR pre-trained on the LVIS dataset \cite{gupta2019lvis} and subsequently fine-tune it to suit our configuration on MSCOCO and VOC. We also conducted an ablation study using their pre-trained model on MSCOCO and observed no significant performance deviation on VOC.

Additionally, we conducted training of \textbf{MD-DETR} on MS-COCO without leveraging any pre-trained model. In other words, we initiated training of \textbf{MD-DETR} from scratch exclusively on $\mathcal{T}_1$ for 50 epochs. Subsequently, we executed our standard fine-tuning process on the remaining tasks ${\mathcal{T}_2 \cdots \mathcal{T}_4}$ for 5 epochs each. While training on $\mathcal{T}_1$ we use a step-scheduler and perform the scheduling at the $40^{th}$ epoch. Moreover, we use multi-scale feature maps extracted from a ResNet-50 \cite{resnet_he2016deep}, pre-trained on ImageNet \cite{deng2009imagenet} (as done in \cite{ddetr}).
As illustrated in Table \ref{tab:ablate_pre}, we observed no significant performance distinction between the model pre-trained on MSCOCO and the one pre-trained solely on $\mathcal{T}_1$. The pre-trained model shows some performance improvement in the initial task $\mathcal{T}_1$, but no/minimal improvement is seen in any subsequent tasks.

\begin{table*}[t]
\centering
\setlength{\tabcolsep}{3pt}
\adjustbox{width=\textwidth}{
\begin{tabular}{@{}l|c|ccc|ccc|ccc@{}}
\toprule
 \textbf{Task IDs} ($\rightarrow$)& \multicolumn{1}{c|}{\textbf{Task 1}} & \multicolumn{3}{c|}{\textbf{Task 2}} & \multicolumn{3}{c|}{\textbf{Task 3}} & \multicolumn{3}{c}{\textbf{Task 4}} \\ \midrule
 

& \begin{tabular}[c]{c}mAP@C\end{tabular} & \begin{tabular}[c]{@{}c}mAP@P\end{tabular} & \begin{tabular}[c]{@{}c}mAP@C\end{tabular} & mAP@A & \begin{tabular}[c]{@{}c@{}}mAP@P\end{tabular} & \begin{tabular}[c]{@{}c@{}}mAP@C\end{tabular} & mAP@A & \begin{tabular}[c]{@{}c@{}}mAP@P\end{tabular} & \begin{tabular}[c]{@{}c@{}}mAP@C\end{tabular} & mAP@A \\ \midrule

COCO+LVIS & 78.1 & 69.1 & 56.5 & 61.2  & 54.6 & 58.3 & 55.4 & 51.5 & 52.7 & 50.2 \\ 
$\mathcal{T}_1$ & 76.2 & 69.0 & 55.9 & 61.1  & 54.5 & 57.7 & 55.3 & 51.4 & 52.1 & 50.0 \\ 

\bottomrule

\end{tabular}%
}
\caption{\textbf{Ablation study on different strategies for pre-training.} We show the effect of using a pre-trained model trained on LVIS+MS-COCO and the one pre-trained solely on $\mathcal{T}_1$. \vspace{-0.3cm}}
\label{tab:ablate_pre}
\end{table*}
\vspace{-0.3cm}

\begin{table}[!t]
\centering
\resizebox{0.6\columnwidth}{!}{%
\begin{tabular}{@{}l|cccc@{}}
\toprule
 & \textbf{Task 1} & \textbf{Task 2} & \textbf{Task 3} & \textbf{Task 4} \\ \midrule
Semantic split & \begin{tabular}[c]{@{}c@{}}Animals, Person, \\ Vehicles\end{tabular} & \begin{tabular}[c]{@{}c@{}}Appliances, Accessories, \\ Outdoor, Furniture\end{tabular} & \begin{tabular}[c]{@{}c@{}}Sports, \\ Food\end{tabular} & \begin{tabular}[c]{@{}c@{}}Electronic, Indoor, \\ Kitchen\end{tabular} \\\midrule
\# training images & 89490 & 55870 & 39402 & 38903 \\
\# test images & 3793 & 2351 & 1642 & 1691 \\
\# train instances & 421243 & 163512 & 114452 & 160794 \\
\# test instances & 17786 & 7159 & 4826 & 7010 \\ \bottomrule
\end{tabular}%
} \vspace{0.8em}
\caption{\textbf{Task composition for MS-COCO given by \cite{OW-DETR}}. The semantic content of each task, along with the distribution of images and instances (objects) across splits, is illustrated. \vspace{-0.2cm}}
\vspace{-0.5em}

\label{tab:data_split}
\end{table}

\begin{table*}[!h]
\centering
\setlength{\tabcolsep}{3pt}
\adjustbox{width=\textwidth}{%
\begin{tabular}{@{}lccccccccccccccccccccc@{}}
\toprule
{\color[HTML]{009901} \textbf{10 + 10 setting}} & aero & cycle & bird & boat & bottle & bus & car & cat & chair & cow & table & dog & horse & bike & person & plant & sheep & sofa & train & tv & mAP \\ \midrule
ILOD \cite{ILOD} & 69.9 & 70.4 & 69.4 & 54.3 & 48 & 68.7 & 78.9 & 68.4 & 45.5 & 58.1 & \cellcolor[HTML]{EEEEEE}59.7 & \cellcolor[HTML]{EEEEEE}72.7 & \cellcolor[HTML]{EEEEEE}73.5 & \cellcolor[HTML]{EEEEEE}73.2 & \cellcolor[HTML]{EEEEEE}66.3 & \cellcolor[HTML]{EEEEEE}29.5 & \cellcolor[HTML]{EEEEEE}63.4 & \cellcolor[HTML]{EEEEEE}61.6 & \cellcolor[HTML]{EEEEEE}69.3 & \cellcolor[HTML]{EEEEEE}62.2 & 63.2 \\

Faster ILOD \cite{peng2020faster} & 72.8 & 75.7 & 71.2 & 60.5 & 61.7 & 70.4 & 83.3 & 76.6 & 53.1 & 72.3 & \cellcolor[HTML]{EEEEEE}36.7 & \cellcolor[HTML]{EEEEEE}70.9 & \cellcolor[HTML]{EEEEEE}66.8 & \cellcolor[HTML]{EEEEEE}67.6 & \cellcolor[HTML]{EEEEEE}66.1 & \cellcolor[HTML]{EEEEEE}24.7 & \cellcolor[HTML]{EEEEEE}63.1 & \cellcolor[HTML]{EEEEEE}48.1 & \cellcolor[HTML]{EEEEEE}57.1 & \cellcolor[HTML]{EEEEEE}43.6 & 62.1 \\ 
ORE $-$ (CC  + EBUI)~\cite{ORE} & 53.3 & 69.2 & 62.4 & 51.8 & 52.9 & 73.6 & 83.7 & 71.7 & 42.8 & 66.8 & \cellcolor[HTML]{EEEEEE}46.8 & \cellcolor[HTML]{EEEEEE}59.9 & \cellcolor[HTML]{EEEEEE}65.5 & \cellcolor[HTML]{EEEEEE}66.1 & \cellcolor[HTML]{EEEEEE}68.6 & \cellcolor[HTML]{EEEEEE}29.8 & \cellcolor[HTML]{EEEEEE}55.1 & \cellcolor[HTML]{EEEEEE}51.6 & \cellcolor[HTML]{EEEEEE}65.3 & \cellcolor[HTML]{EEEEEE}51.5 & 59.4 \\
ORE $-$ EBUI~\cite{ORE} & 63.5 & 70.9 & 58.9 & 42.9 & 34.1 & 76.2 & 80.7 & 76.3 & 34.1 & 66.1 & \cellcolor[HTML]{EEEEEE}56.1 & \cellcolor[HTML]{EEEEEE}70.4 & \cellcolor[HTML]{EEEEEE}80.2 & \cellcolor[HTML]{EEEEEE}72.3 & \cellcolor[HTML]{EEEEEE}81.8 & \cellcolor[HTML]{EEEEEE}42.7 & \cellcolor[HTML]{EEEEEE}71.6 & \cellcolor[HTML]{EEEEEE}68.1 & \cellcolor[HTML]{EEEEEE}77 & \cellcolor[HTML]{EEEEEE}67.7 & 64.5 \\ 
OW-DETR\cite{OW-DETR} & 61.8 & 69.1 & 67.8 & 45.8 & 47.3 & 78.3 & 78.4 & 78.6 & 36.2 & 71.5 &  \cellcolor[HTML]{EEEEEE} 57.5 &  \cellcolor[HTML]{EEEEEE} 75.3 &  \cellcolor[HTML]{EEEEEE} 76.2 &  \cellcolor[HTML]{EEEEEE} 77.4 &  \cellcolor[HTML]{EEEEEE} 79.5 &  \cellcolor[HTML]{EEEEEE} 40.1 &  \cellcolor[HTML]{EEEEEE} 66.8 &  \cellcolor[HTML]{EEEEEE} 66.3 &  \cellcolor[HTML]{EEEEEE} 75.6 & \cellcolor[HTML]{EEEEEE} 64.1 & 65.7 \\ 

PROB \cite{prob_zohar2023prob} & 70.4 & 75.4 & 67.3 & 48.1 & 55.9 & 73.5 & 78.5 & 75.4 & 42.8 & 72.2&  \cellcolor[HTML]{EEEEEE} 64.2 &  \cellcolor[HTML]{EEEEEE} 73.8 &  \cellcolor[HTML]{EEEEEE} 76.0 &  \cellcolor[HTML]{EEEEEE} 74.8 &  \cellcolor[HTML]{EEEEEE} 75.3 &  \cellcolor[HTML]{EEEEEE} 40.2 &  \cellcolor[HTML]{EEEEEE} 66.2 &  \cellcolor[HTML]{EEEEEE} 73.3 &  \cellcolor[HTML]{EEEEEE} 64.4 & \cellcolor[HTML]{EEEEEE} 64.0 & 66.5 \\ 

\midrule

\textbf{MD-DETR} & 72.4 & 76.7 & 72.1 & 62.5 & 59.2 & 78.7 & 84.1 & 76.2 & 56.8 & 73.5 &  \cellcolor[HTML]{EEEEEE} 67.5 &  \cellcolor[HTML]{EEEEEE} 74.2 &  \cellcolor[HTML]{EEEEEE} 79.3 &  \cellcolor[HTML]{EEEEEE} 80.1 &  \cellcolor[HTML]{EEEEEE} 82.3 &  \cellcolor[HTML]{EEEEEE} 49.2 &  \cellcolor[HTML]{EEEEEE} 68.5 &  \cellcolor[HTML]{EEEEEE} 77.4 &  \cellcolor[HTML]{EEEEEE} 75.9 & \cellcolor[HTML]{EEEEEE} 68.2 & \textbf{73.2} \\ 

\midrule\midrule

{\color[HTML]{009901} \textbf{15 + 5 setting}} & aero & cycle & bird & boat & bottle & bus & car & cat & chair & cow & table & dog & horse & bike & person & plant & sheep & sofa & train & tv & mAP \\ \midrule
ILOD \cite{ILOD} & 70.5 & 79.2 & 68.8 & 59.1 & 53.2 & 75.4 & 79.4 & 78.8 & 46.6 & 59.4 & 59 & 75.8 & 71.8 & 78.6 & 69.6 & \cellcolor[HTML]{EEEEEE}33.7 & \cellcolor[HTML]{EEEEEE}61.5 & \cellcolor[HTML]{EEEEEE}63.1 & \cellcolor[HTML]{EEEEEE}71.7 & \cellcolor[HTML]{EEEEEE}62.2 & 65.8 \\

Faster ILOD \cite{peng2020faster} & 66.5 & 78.1 & 71.8 & 54.6 & 61.4 & 68.4 & 82.6 & 82.7 & 52.1 & 74.3 & 63.1 & 78.6 & 80.5 & 78.4 & 80.4 & \cellcolor[HTML]{EEEEEE}36.7 & \cellcolor[HTML]{EEEEEE}61.7 & \cellcolor[HTML]{EEEEEE}59.3 & \cellcolor[HTML]{EEEEEE}67.9 & \cellcolor[HTML]{EEEEEE}59.1 & 67.9 \\ 
ORE $-$ (CC  + EBUI)~\cite{ORE} & 65.1 & 74.6 & 57.9 & 39.5 & 36.7 & 75.1 & 80 & 73.3 & 37.1 & 69.8 & 48.8 & 69 & 77.5 & 72.8 & 76.5 & \cellcolor[HTML]{EEEEEE}34.4 & \cellcolor[HTML]{EEEEEE}62.6 & \cellcolor[HTML]{EEEEEE}56.5 & \cellcolor[HTML]{EEEEEE}80.3 & \cellcolor[HTML]{EEEEEE}65.7 & 62.6 \\
ORE $-$ EBUI~\cite{ORE} & 75.4 & 81 & 67.1 & 51.9 & 55.7 & 77.2 & 85.6 & 81.7 & 46.1 & 76.2 & 55.4 & 76.7 & 86.2 & 78.5 & 82.1 & \cellcolor[HTML]{EEEEEE}32.8 & \cellcolor[HTML]{EEEEEE}63.6 & \cellcolor[HTML]{EEEEEE}54.7 & \cellcolor[HTML]{EEEEEE}77.7 & \cellcolor[HTML]{EEEEEE}64.6 & 68.5 \\ 
OW-DETR \cite{OW-DETR}& 77.1 & 76.5 & 69.2 & 51.3 & 61.3 & 79.8 & 84.2 & 81.0 & 49.7 & 79.6 & 58.1 & 79.0 & 83.1 & 67.8 & 85.4 & \cellcolor[HTML]{EEEEEE}33.2 & \cellcolor[HTML]{EEEEEE}65.1 & \cellcolor[HTML]{EEEEEE}62.0 & \cellcolor[HTML]{EEEEEE}73.9 & \cellcolor[HTML]{EEEEEE}65.0 & 69.4 \\ 
PROB \cite{prob_zohar2023prob} & 77.9 &77.0 &77.5 &56.7 &63.9 &75.0 &85.5 &82.3 &50.0 &78.5 &63.1 &75.8 &80.0 &78.3 &77.2 &  \cellcolor[HTML]{EEEEEE} 38.4 &  \cellcolor[HTML]{EEEEEE} 69.8 &  \cellcolor[HTML]{EEEEEE} 57.1 &  \cellcolor[HTML]{EEEEEE} 73.7 & \cellcolor[HTML]{EEEEEE} 64.9 & 70.1 \\\midrule

\textbf{MD-DETR} & 80.1 & 82.4 & 78.5 & 60.7 & 65.4 & 82.1 & 86.1 & 82.5 & 55.7 & 80.7 & 63.9 & 79.3 & 83.4 & 79.5 & 83.5 &  \cellcolor[HTML]{EEEEEE} 42.7 &  \cellcolor[HTML]{EEEEEE} 72.3 &  \cellcolor[HTML]{EEEEEE} 64.2 &  \cellcolor[HTML]{EEEEEE} 80.1 & \cellcolor[HTML]{EEEEEE}69.3 & \textbf{76.7} \\

\midrule\midrule

{\color[HTML]{009901} \textbf{19 + 1 setting}} & aero & cycle & bird & boat & bottle & bus & car & cat & chair & cow & table & dog & horse & bike & person & plant & sheep & sofa & train & tv & mAP \\ \midrule
ILOD \cite{ILOD} & 69.4 & 79.3 & 69.5 & 57.4 & 45.4 & 78.4 & 79.1 & 80.5 & 45.7 & 76.3 & 64.8 & 77.2 & 80.8 & 77.5 & 70.1 & 42.3 & 67.5 & 64.4 & 76.7 & \cellcolor[HTML]{EEEEEE}62.7 & 68.2 \\

Faster ILOD \cite{peng2020faster} & 64.2 & 74.7 & 73.2 & 55.5 & 53.7 & 70.8 & 82.9 & 82.6 & 51.6 & 79.7 & 58.7 & 78.8 & 81.8 & 75.3 & 77.4 & 43.1 & 73.8 & 61.7 & 69.8 & \cellcolor[HTML]{EEEEEE}61.1 & 68.5 \\
ORE $-$ (CC  + EBUI)~\cite{ORE} & 60.7 & 78.6 & 61.8 & 45 & 43.2 & 75.1 & 82.5 & 75.5 & 42.4 & 75.1 & 56.7 & 72.9 & 80.8 & 75.4 & 77.7 & 37.8 & 72.3 & 64.5 & 70.7 & \cellcolor[HTML]{EEEEEE}49.9 & 64.9 \\
ORE $-$ EBUI~\cite{ORE} & 67.3 & 76.8 & 60 & 48.4 & 58.8 & 81.1 & 86.5 & 75.8 & 41.5 & 79.6 & 54.6 & 72.8 & 85.9 & 81.7 & 82.4 & 44.8 & 75.8 & 68.2 & 75.7 & \cellcolor[HTML]{EEEEEE}60.1 & 68.8 \\ 
OW-DETR \cite{OW-DETR} & 70.5 & 77.2 & 73.8 & 54.0 & 55.6 & 79.0 & 80.8 & 80.6 & 43.2 & 80.4 & 53.5 & 77.5 & 89.5 & 82.0 & 74.7 & 43.3 & 71.9 & 66.6 & 79.4 & \cellcolor[HTML]{EEEEEE}62.0 & 70.2 \\
PROB \cite{prob_zohar2023prob} & 80.3 &78.9 &77.6 &59.7 &63.7 &75.2 &86.0 &83.9 &53.7 &82.8 &66.5 &82.7 &80.6 &83.8 &77.9 &48.9 &74.5 &69.9 &77.6  &\cellcolor[HTML]{EEEEEE}48.5 & 72.6 \\ 

\midrule

\textbf{MD-DETR} & 82.4 & 78.2 & 77.9 & 60.5 & 66.4 & 79.8 & 86.8 & 84.7 & 55.3 & 83.5 & 67.7 & 82.5 & 83.4 & 83.9 & 78.3 & 54.3 & 75.1 & 71.1 & 79.2  &\cellcolor[HTML]{EEEEEE} 67.2 &   \textbf{76.1} \\

\bottomrule
\end{tabular}%
}\vspace{0.8em}
\caption{\textbf{State-of-the-art comparison for continual object detection on PASCAL-VOC.} We show 3 different A+B settings; where task B is shaded in gray. \vspace{-0.3cm}
}
\label{tab:sup_pascal}
\end{table*}

\begin{table}[!b]
\caption{Lookup table for notations in the paper.}
\vspace{-0.9cm}
\label{tab:notations_lookup}
\begin{center}
\begin{tabular}{l@{\hskip 0.4in}l}
\toprule
Notation\          & Description \\
\hline
$\mathcal{T}_i$    & task/step $i$ \\
$C^{\mathcal{T}_i}$    & object labels/classes present in task/step $i$ \\
$\emptyset$ & null set \\
$\theta^{\mathcal{T}_t}$ & parameters of the model at task/step $t$ \\
$\theta_\nabla$ & frozen parameters of the model: encoder and decoder \\
$\theta^{\mathcal{T}_t*}$ & learnable parameters at task/step $t$: memory units, class and bbox embed \\

$\mathbf{M}$ & memory modules \\
$L_m$ & length of memory modules \\
$N_m$ & number of memory modules \\
$D$ & dimension of embedding \\
$\mathbf{K}$ & key vector of the memory units \\
$\rho$ & similarity function: cosine \\
$Q$ & query retrieval function for memories \\
$\phi$ & deterministic feature extractor \\
$\mathbf{A}$ & attention matrix for memory units \\
$\textit{w}$ & weight vector for each memory module \\
$g_\psi$ & ranking function parametrized by $\psi$ \\
$\alpha$ & ranking weights produced by $g_\psi$\\
$\hat{\sigma}$ & optimal assignment from Hungarian matching algorithm \\
$\delta_{bt}$ & user-defined threshold for background thresholding \\
$\lambda_Q$ & coefficient for query-retrieval regularization \\
$mAP$ & mean-average precision at intersection-over-union=0.50\\
$mAP@P$ & $mAP$ for previously seen classes \\
$mAP@C$ & $mAP$ for previously current classes \\
$mAP@A$ & $mAP$ for all seen classes till now \\

$m$                & index for attention head \\
$l$                & index for feature level of key element \\
$q$                & index for query element \\
$k$                & index for key element \\
$N_q$              & number of query elements \\
$N_k$              & number of key elements \\
$L$                & number of input feature levels \\
$H$                & height of input feature map \\
$W$                & width of input feature map \\
$\textit{A}_{mqk}$       & attention weight of $q^{th}$ query to $k^{th}$ key at $m^{th}$ head \\
$\textit{z}_q$            & input feature of $q^{th}$ query/proposal \\
$\textit{p}_q$            & 2-d coordinate of reference point for $q^{th}$ query \\
$\textbf{W}_m$            & output projection matrix at $m^{th}$ head \\
$\textbf{U}_m$            & input query projection matrix at $m^{th}$ head \\
$\textbf{V}_m$            & input key projection matrix at $m^{th}$ head \\
$\textbf{W}'_m$           & input value projection matrix at $m^{th}$ head \\

\bottomrule

\end{tabular}
\end{center}
\end{table}

\section{Continual split for MS-COCO given by \cite{OW-DETR}}

In this work, we use the semantic split of MS-COCO into 4 multi-step continual tasks, as defined by \cite{OW-DETR}. We show the distribution of images and objects across these splits in Table \ref{tab:data_split}.

\section{Per-class Evaluation on PASCAL-VOC}
We evaluate \textbf{MD-DETR} on PASCAL-VOC (2007) using average-precision $aP$ for individual classes and aggregate the $aP$ values across to get $mAP$. The results are shown in Table \ref{tab:sup_pascal}.

 \section{Notations}
 All notations used in this work are listed in Table \ref{tab:notations_lookup}.

\section{Time-space analysis}
\begin{figure}[!h]
    \vspace{-0.15in}
  \centering
  \includegraphics[width=\linewidth]{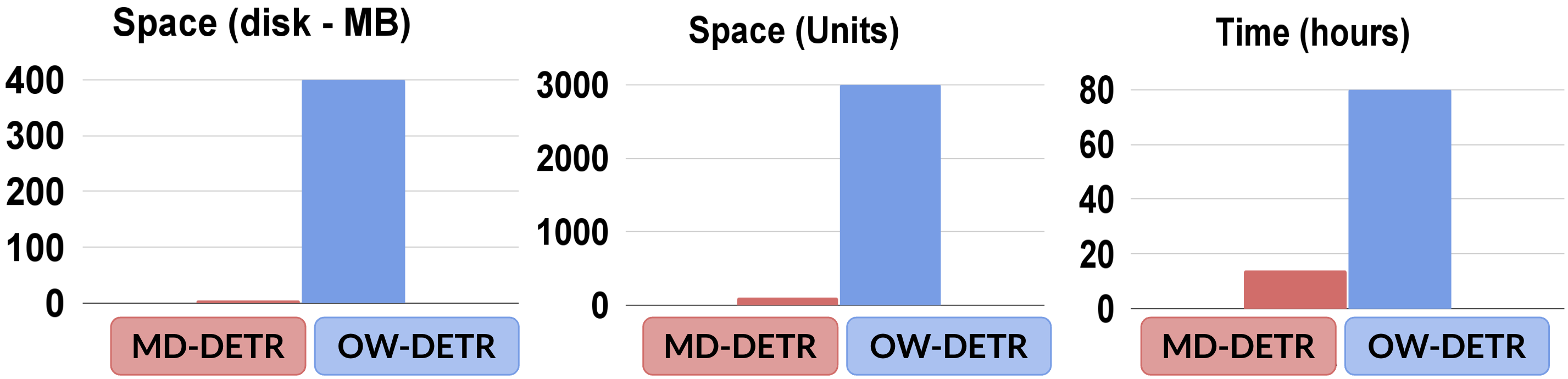}
  \vspace{-0.7cm}
   \caption{Time-space comparison b/w MD-DETR \& OW-DETR.}
   \vspace{-0.3cm}
   \label{fig:compare}
\end{figure}

In Figure \ref{fig:compare}, we present a time-space analysis of MD-DETR and compare it to OW-DETR \cite{OW-DETR}, a SoTA replay-buffer method. For the memory/space analysis, we consider two aspects: 1) disk space requirement (measured in megabytes - MB), and 2) the number of units required. For MD-DETR, units refer to the number of memory elements, whereas for OW-DETR, units indicate the total size of the replay buffer (samples stored in the buffer). The time analysis, conducted on the MS-COCO dataset, reflects the training time required in hours.

\section{Qualitative analysis of Background Relegation}

\begin{figure*}[!b]
        \centering
        \includegraphics[scale=0.22]{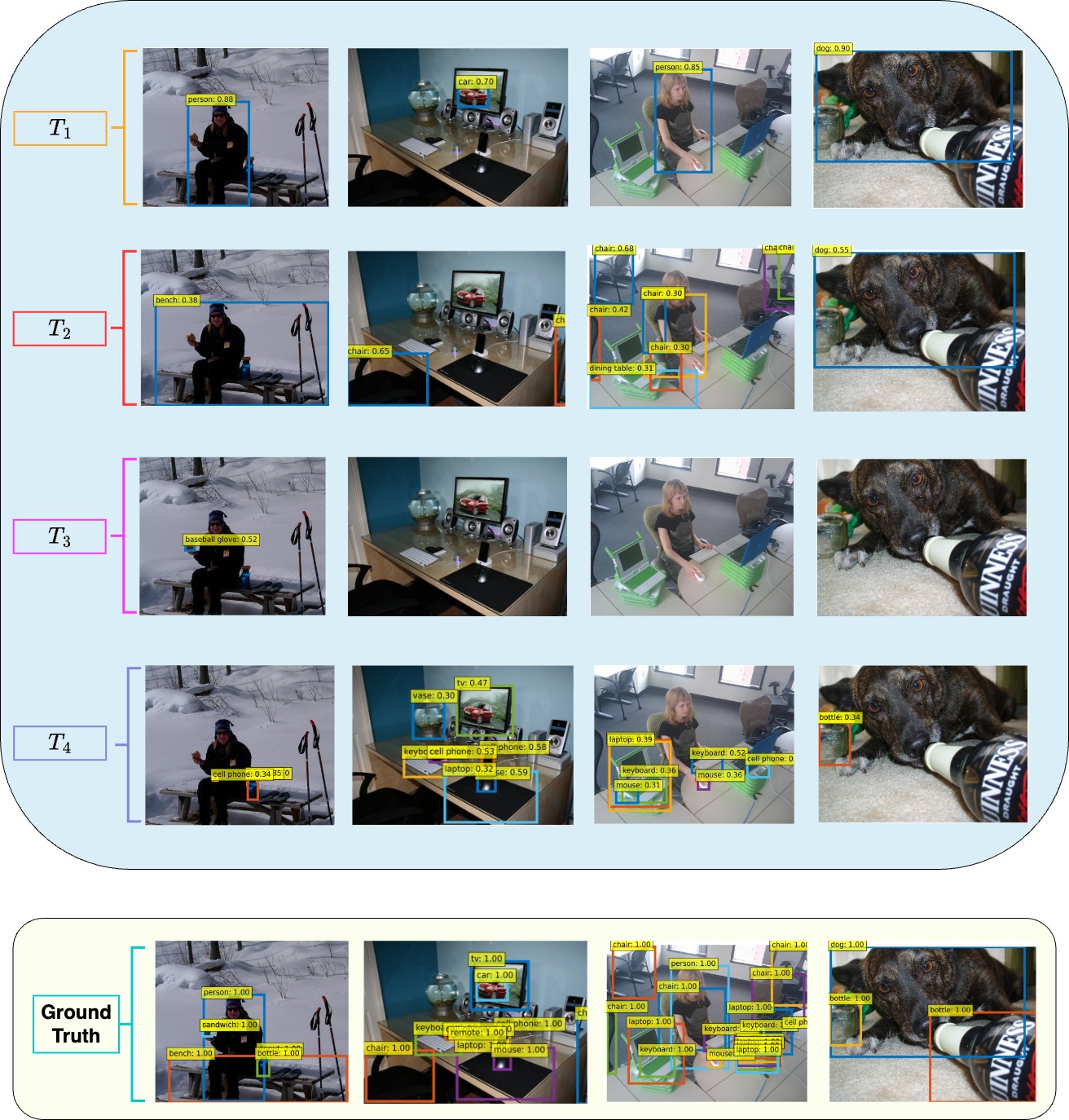}
        \caption{{\bf Progression of MD-DETR - BT performance from $\mathcal{T}_1 \cdots \mathcal{T}_4$ when trained in a multi-step class incremental setting on MS-COCO.} In the images shown in the first column (1$^{st}$ image for each block), the class {\em \{person\}} is relegated by \textbf{MD-DETR} - \textbf{BT}; similarly, for the second column, the category {\em \{car\}} is relegated, the third column displays relegation of classes {\em \{person, chair\}}, and the fourth column shows the relegation of the category {\em \{dog\}}. The ground truth block shows images with all annotations across all the four tasks $\{\mathcal{T}_1 \cdots \mathcal{T}_4\}$.
        }
        \label{fig:sup_viz_nobt}
        \vspace{-4pt}
\end{figure*}

\begin{figure*}[!t]
        \centering
        \includegraphics[scale=0.24]{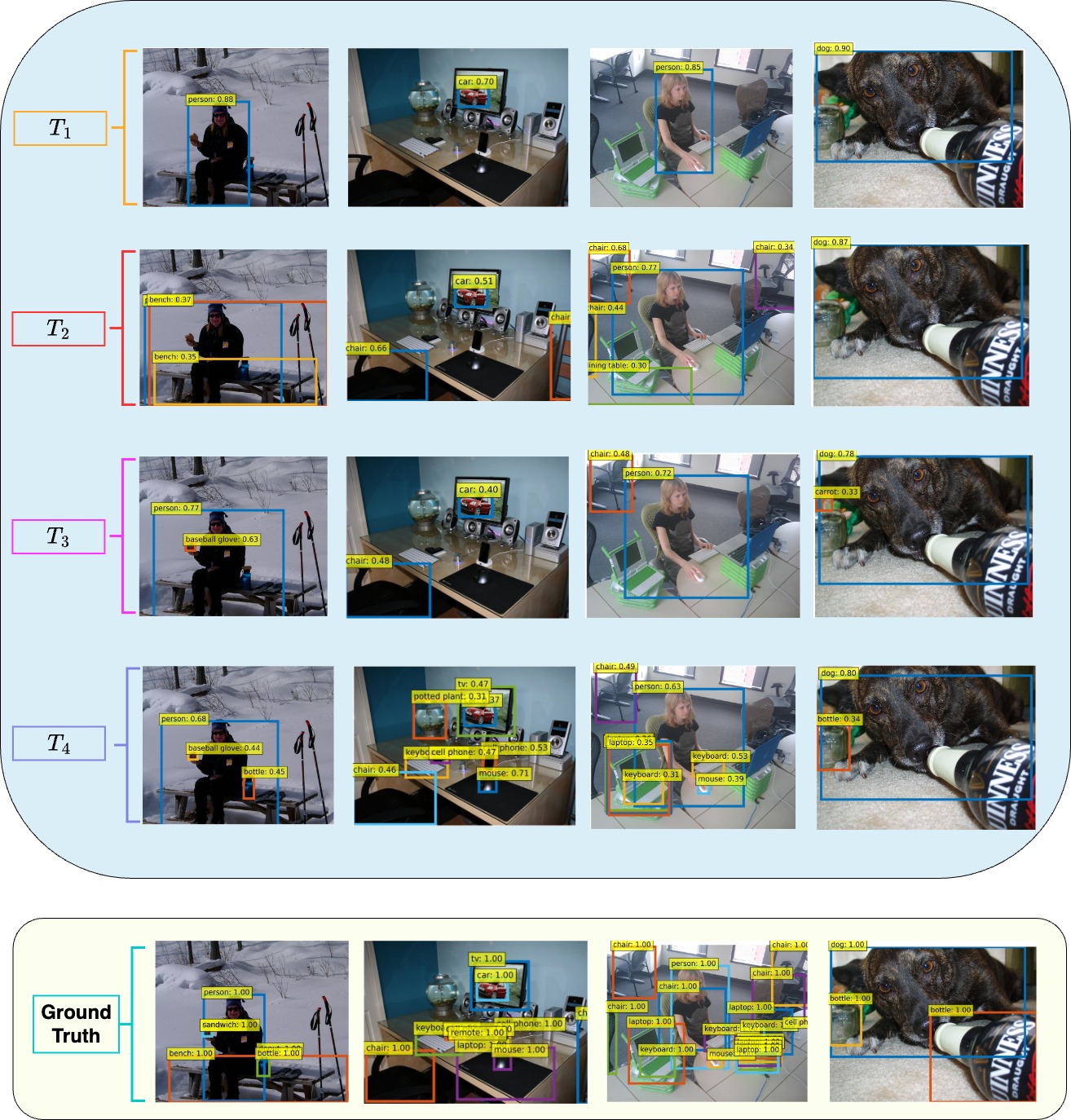}
        \caption{{\bf Progression of MD-DETR + BT performance from $\mathcal{T}_1 \cdots \mathcal{T}_4$ when trained in a multi-step class incremental setting on MS-COCO.} \textbf{MD-DETR + BT} successfully remembers most objects from the past due to the minimization of background relegation.  The ground truth block shows images with all annotations across all the four tasks $\{\mathcal{T}_1 \cdots \mathcal{T}_4\}$.
        }
        \label{fig:sup_viz_bt}
        \vspace{-4pt}
\end{figure*}

Following Section 5.3 (main draft), we conduct a detailed qualitative analysis of the performance of \textbf{MD-DETR}. We emphasize the issue of background relegation across all four tasks on MS-COCO and illustrate \textbf{MD-DETR}'s effectiveness in mitigating the relegation of previously detected objects. Initially, we show the qualitative outcomes of \textbf{MD-DETR - BT} (without background thresholding) in Figure \ref{fig:sup_viz_nobt}. This configuration struggles to retain most objects from $\mathcal{T}_3$, primarily due to the relegation of these object categories. Conversely, the \textbf{MD-DETR + BT} architecture effectively retains most previously detected objects, as evidenced in Figure \ref{fig:sup_viz_bt}.

\clearpage

%
%
\bibliographystyle{splncs04}
\bibliography{main}
\end{document}